\documentclass[10pt,twocolumn,letterpaper]{article}

\usepackage{subcaption}

\usepackage{iccv}
\usepackage{times}
\usepackage{epsfig}
\usepackage{graphicx}
\usepackage{amsmath}
\usepackage{amssymb}

\usepackage[accsupp]{axessibility}

\usepackage{bm}
\usepackage{booktabs}
\usepackage{multirow}
\usepackage[table,xcdraw]{xcolor}
\usepackage{mathtools}
\usepackage{bbm}
\usepackage{nicefrac}
\DeclarePairedDelimiter\abs{\lvert}{\rvert}

\newcommand{\fraccomma}{\genfrac{}{}{0pt}{}{}{,}}
\newcommand{\defeq}{\vcentcolon=}

\graphicspath{ {./images/} }

\makeatletter
\def\ps@myheadings{%
    \let\@oddfoot\@empty\let\@evenfoot\@empty
    \def\@evenhead{\thepage\hfil\slshape\leftmark}%
    \def\@oddhead{{\slshape\rightmark}\hfil\thepage}%
    \let\@mkboth\@gobbletwo
    \let\sectionmark\@gobble
    \let\subsectionmark\@gobble
    }
  \if@titlepage
  \renewcommand\maketitle{\begin{titlepage}%
  \let\footnotesize\small
  \let\footnoterule\relax
  \let \footnote \thanks
  \null\vfil
  \vskip 60\p@
  \begin{center}%
    {\LARGE \@title \par}%
    \vskip 3em%
    {\large
     \lineskip .75em%
      \begin{tabular}[t]{c}%
        \@author
      \end{tabular}\par}%
      \vskip 1.5em%
    {\large \@date \par}%       % Set date in \large size.
  \end{center}\par
  \@thanks
  \vfil\null
  \end{titlepage}%
  \setcounter{footnote}{0}%
}
\else
\renewcommand\maketitle{\par
  \begingroup
    \renewcommand\thefootnote{\@fnsymbol\c@footnote}%
    \def\@makefnmark{\rlap{\@textsuperscript{\normalfont\@thefnmark}}}%
    \long\def\@makefntext##1{\parindent 1em\noindent
            \hb@xt@1.8em{%
                \hss\@textsuperscript{\normalfont\@thefnmark}}##1}%
    \if@twocolumn
      \ifnum \col@number=\@ne
        \@maketitle
      \else
        \twocolumn[\@maketitle]%
      \fi
    \else
      \newpage
      \global\@topnum\z@   % Prevents figures from going at top of page.
      \@maketitle
    \fi
    \thispagestyle{plain}\@thanks
  \endgroup
  \setcounter{footnote}{0}%
}
\makeatother

\usepackage[pagebackref=true,breaklinks=true,letterpaper=true,colorlinks,bookmarks=false]{hyperref}

\iccvfinalcopy % *** Uncomment this line for the final submission

\def\iccvPaperID{****} % *** Enter the ICCV Paper ID here

\ificcvfinal\fi

\begin{document}

%%%%%%%%% TITLE
\title{Robust \textit{e}-NeRF: NeRF from Sparse \& Noisy Events under Non-Uniform Motion}

\author{Weng Fei Low \qquad Gim Hee Lee\\
The NUS Graduate School's Integrative Sciences and Engineering Programme (ISEP)\\
Institute of Data Science (IDS), National University of Singapore\\
Department of Computer Science, National University of Singapore\\
{\tt\small \{wengfei.low, gimhee.lee\}@comp.nus.edu.sg}\\
\small{\url{https://wengflow.github.io/robust-e-nerf}}
% \\
}

\maketitle
\ificcvfinal\thispagestyle{empty}\fi

%%%%%%%%% ABSTRACT
\begin{abstract}

Event cameras offer many advantages over standard cameras due to their distinctive principle of operation: low power, low latency, high temporal resolution and high dynamic range. Nonetheless, the success of many downstream visual applications also hinges on an efficient and effective scene representation, where Neural Radiance Field (NeRF) is seen as the leading candidate. Such promise and potential of event cameras and NeRF inspired recent works to investigate on the reconstruction of NeRF from moving event cameras. However, these works are mainly limited in terms of the dependence on dense and low-noise event streams, as well as generalization to arbitrary contrast threshold values and camera speed profiles. In this work, we propose Robust \textit{e}-NeRF, a novel method to directly and robustly reconstruct NeRFs from moving event cameras under various real-world conditions, especially from sparse and noisy events generated under non-uniform motion. It consists of two key components: a realistic event generation model that accounts for various intrinsic parameters (\eg time-independent, asymmetric threshold and refractory period) and non-idealities (\eg pixel-to-pixel threshold variation), as well as a complementary pair of normalized reconstruction losses that can effectively generalize to arbitrary speed profiles and intrinsic parameter values without such prior knowledge. Experiments on real and novel realistically simulated sequences verify our effectiveness. Our code, synthetic dataset and improved event simulator are public.

\end{abstract}

\begin{figure}[t]
    \begin{center}
    \includegraphics[width=1\linewidth,trim={1.1cm 0.5cm 3.1cm 3.6cm},clip]{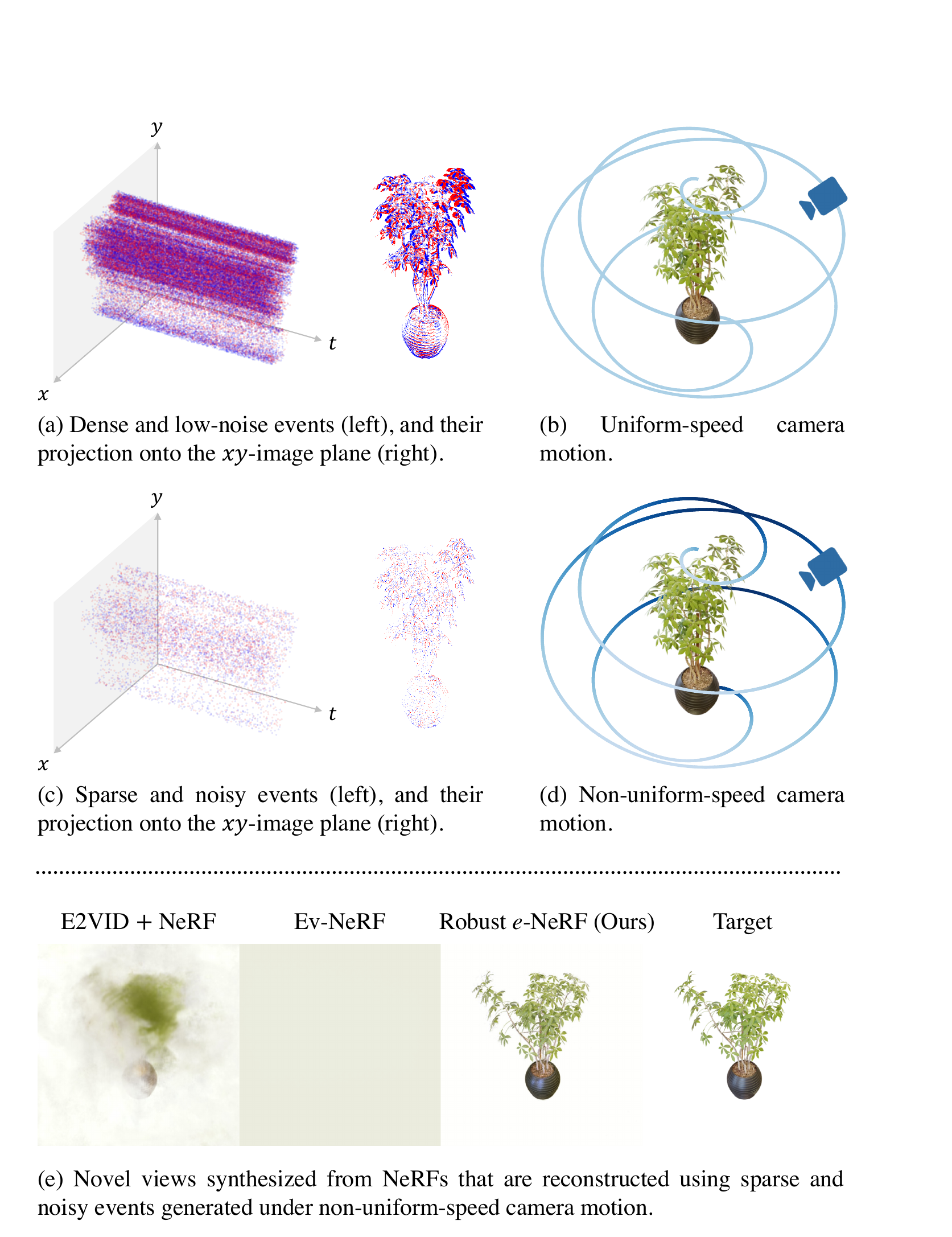}
    \end{center}
    % \vspace{-0.75em}
    \caption{Existing works on NeRF reconstruction from moving event cameras heavily rely on (a) temporally dense and low-noise events generated under roughly (b) uniform-speed camera motion. In contrast, our method, Robust \textit{e}-NeRF is able to directly and robustly reconstruct NeRFs from (c) sparse and noisy events generated under (d) non-uniform camera motion, as shown in (e).}
    \vspace{-0.65em}
    \label{fig:teaser}
\end{figure}

%%%%%%%%% BODY TEXT
\vspace{-0.65em}
\section{Introduction}
\label{sec:intro}

Event cameras are bio-inspired sensors that represent a paradigm shift in visual acquisition and processing. This is attributed to its fundamentally distinctive principle of operation, where its pixels independently respond to log-intensity changes in an asynchronous manner, yielding a stream of \textit{events}, rather than measuring absolute linear intensity synchronously at a constant rate, as done in standard cameras. Such unique properties contribute to their multitude of advantages over standard cameras \cite{gallego2020_event_survey}: \emph{low power}, \emph{low latency}, \emph{high temporal resolution} and \emph{high dynamic range}, thereby the recent success of event-based \cite{kim2016, rebecq2016_evo, gallego2018_contrastmax, falanga2020_quadrotor, rebecq2021_e2vid} or event-image hybrid  \cite{vidal2018_ultimateslam, tulyakov2021_timelens, hidalgo2022_dso} applications.

The success of many downstream visual applications in robotics, computer vision, graphics and virtual/augmented reality also hinges on an efficient and effective representation to encode various information of the scene being interacted with. Neural scene representations \cite{mildenhall2020_nerf, oechsle2021_unisurf, wang2021_neus, yariv2021_volsdf, low2022_mna}, especially neural fields \cite{xie2022_fields_survey}, have recently emerged as promising candidates for future applications, owing to their continuous nature and memory efficiency. This trend is further driven by the exceptional capabilities and photo-realism of \textit{Neural Radiance Field} (NeRF) \cite{mildenhall2020_nerf}-based works \cite{wang2021_neus, mildenhall2022_rawnerf, sucar2021_imap, park2021_hypernerf, peng2021_animatablenerf, poole2023_dreamfusion}.

Motivated by such promise and potential of event cameras and NeRF, we are interested in studying the following research question: \emph{How to robustly reconstruct a NeRF from a moving event camera under general real-world conditions?} One simple way is to retrofit an events-to-video reconstruction method \cite{rebecq2021_e2vid, stoffregen2020_e2vid+, paredes2021_ssl_e2vid} to NeRF. However, such a naïve approach is inherently limited by the low photometric accuracy and consistency of the reconstructed video frames, since they are assumed to be true observations of the scene.

On the contrary, recent works, such as EventNeRF \cite{rudnev2022_eventnerf}, Ev-NeRF \cite{hwang2022_evnerf} and E-NeRF \cite{klenk2022_enerf}, have proposed to reconstruct NeRFs directly using events via alternative reconstruction losses inspired or derived from an event generation model. Nonetheless, these works heavily rely on a temporally dense and low-noise event stream, which is generally inaccessible in practice due to the presence of \textit{refractory period} (\ie pixel dead-time after generating an event) and pixel-to-pixel variation in the \textit{contrast threshold} (\ie minimum log-intensity change for event generation). Such a limitation can be partly attributed to the accumulation of successive events at each pixel over time intervals, as performed in these works. Moreover, the reduction in contrast sensitivity resulting from the event accumulation also leads to a loss of detail in the reconstruction.

In addition, these methods do not directly and effectively generalize to arbitrary contrast threshold values and camera speed profiles, as their optimal hyper-parameter configuration greatly depends on the contrast threshold and speed of motion. EventNeRF and E-NeRF assumes symmetric positive and negative thresholds, which generally does not hold true in practice. While joint optimization of the contrast threshold is supported in Ev-NeRF, an additional regularization is necessary to prevent degeneracy. Furthermore, the assumption of time-varying thresholds made in Ev-NeRF and E-NeRF, which is not well supported by the literature, also leads to a reduction in reconstruction accuracy as shown in our experiments.

\vspace{-0.65em}
\paragraph{Contributions.}

We propose Robust \textit{e}-NeRF, a novel method to directly and robustly reconstruct NeRFs from moving event cameras under various real-world conditions, especially from temporally sparse and noisy event streams given by event cameras in non-uniform motion.

In particular, we incorporate a more realistic event generation model that accounts for various intrinsic parameters (\eg time-independent, asymmetric contrast threshold and refractory period) and non-idealities (\eg pixel-to-pixel threshold variation). Furthermore, we introduce two complementary normalized reconstruction losses that are not only effectively invariant to the camera motion speed and threshold scale, but also minimally influenced by asymmetric thresholds. This allows for their effective generalization, as well as the regularization-free joint optimization of unknown contrast threshold and refractory period from poor initializations. The first loss serves as the primary loss for high-fidelity reconstruction, while the second acts as a smoothness constraint for better regularization of textureless regions. As both loss functions do not involve event accumulation, detailed and robust reconstruction from sparse and noisy events can be achieved. Our experiments on novel sequences, simulated using an improved version of ESIM \cite{rebecq2018_esim}, and real sequences from TUM-VIE \cite{klenk2021_tumvie} verify the effectiveness of Robust \textit{e}-NeRF. We publicly release our code, synthetic event dataset and improved ESIM.

\section{Related Work}
\label{sec:related}

\paragraph{Event-based Scene Reconstruction.}

Successful reconstruction of geometry, and possibly appearance, from events has been demonstrated in notable works, such as \cite{rebecq2018_emvs, gallego2018_contrastmax, zhou2018_stereo_evrec, rebecq2016_evo, kim2016, zhou2021_stereovo}. Nonetheless, reconstruction of scene appearance is largely limited to diffuse surfaces due to the adoption of Lambertian surface assumption via brightness constancy \cite{rebecq2016_evo, kim2016}. Moreover, existing methods can mainly recover semi-dense geometry, in the form of discrete depth maps or point clouds, corresponding to edges (more precisely, locations with high perceived spatial intensity gradient) in the scene. This is due to the fact that events are primarily generated along edges under relative motion \cite{gallego2020_event_survey}.

While \cite{xiao2022_event_dense_rec} achieved dense diffuse reconstruction by applying the classic \textit{Structure-from-Motion} (SfM) and \textit{Multi-View Stereo} (MVS) pipelines on video frames reconstructed from events \cite{rebecq2021_e2vid}, its performance is intrinsically limited by the accuracy of the recovered frames, as similarly discussed in \autoref{sec:intro}. Although limited to simple, object-level mesh or specialized parametric models, \cite{rudnev2021_eventhands, nehvi2021_diff_event, xue2022_contours} also demonstrated dense diffuse reconstruction via \textit{Analysis-by-Synthesis} or equivalently \textit{Vision-as-Inverse-Graphics} \cite{kato2020_dr_survey}. In contrast, we aim to reconstruct \emph{dense, continuous} scene geometry and \emph{view-dependent} appearance, in the form of a NeRF, \emph{directly} from the raw event stream.

\vspace{-0.65em}
\paragraph{Reconstructing Neural Radiance Fields.}

In general, visual reconstruction of neural scene representations, including NeRFs, is achieved via \textit{Analysis-by-Synthesis}. Nonetheless, NeRF derivatives mainly focus on the reconstruction from dense \cite{martin2021_nerfw, barron2021_mipnerf, barron2022_mipnerf360} or sparse \cite{niemeyer2022_regnerf, kim2022_infonerf} multi-view images, possibly with depth maps \cite{attal2021_torf, azinovic2022_rgbdnerf} or point clouds \cite{roessle2022_dense_depth_nerf, deng2022_dsnerf, rematas2022_urban}.

The reconstruction of NeRFs from events was first proposed and investigated in Ev-NeRF \cite{hwang2022_evnerf}, E-NeRF \cite{klenk2022_enerf} and EventNeRF \cite{rudnev2022_eventnerf}. E-NeRF also explored on using a combination of events and images for NeRF reconstruction. However, these works suffer from various limitations, as outlined in \autoref{sec:intro}. Moreover, EventNeRF also relies on the access to the analytic camera trajectory, in contrast to E-NeRF and our work which only require constant-rate camera poses. Furthermore, inconsistent sets of loss functions and hyper-parameters were also adopted across different scenes in E-NeRF. In addition, E-NeRF with normalized and no-event losses also requires the contrast threshold to be known as \textit{a priori}, which is hard to achieve in practice.

\section{Our Method}
\label{sec:method}

We first briefly introduce the \textit{Neural Radiance Field} (NeRF) scene representation (\autoref{subsec:method:nerf}) . Next, we detail the event generation model (\autoref{subsec:method:egm}) and normalized training losses (\autoref{subsec:method:training}) proposed to robustly reconstruct NeRFs from event cameras. Lastly, we describe a \textit{Gamma Correction}-based approach to align the radiance levels of the synthesized views to a set of given reference views (\autoref{subsec:method:gamma}).

\subsection{Preliminaries: Neural Radiance Fields}
\label{subsec:method:nerf}

{Neural Radiance Field} (NeRF) \cite{mildenhall2020_nerf} represents a scene using a \textit{Multi-Layer Perceptron} (MLP) $F_\Theta : (\bm{x}, \bm{d}) \mapsto (\bm{c}, \sigma)$ that maps 3D position $\bm{x} = (x, y, z)$ and 2D viewing direction $\bm{d} = (\theta, \phi)$ to its corresponding directional emitted radiance, or simply color, $\bm{c} = (r, g, b)$ and volume density $\sigma$. From this representation, the estimated incident radiance $\hat{\bm{L}}$ at a given pixel $\bm{u}$ can be computed using the volume rendering equation with quadrature \cite{tagliasacchi2022_nerf_digest}, as follows:
\begin{equation}
    \begin{split}
    	\hat{\bm{L}}(\bm{u}) &= \sum_{i=1}^{N} T_i (1 - \exp(-\sigma_i \delta_i)) \bm{c}_i \ ,\\
        T_i &= \exp(-\sum_{j=1}^{i-1} \sigma_j \delta_j) \ ,
    \end{split}
    \label{eq:L-hat}
\end{equation}
where $\sigma_i$ and $\bm{c}_i$ are the volume density and emitted radiance, respectively, of a sample $\bm{x}_i$ along the back-projected ray through the pixel, which has direction $\bm{d}$ from the camera center $\bm{o}$. The sample $\bm{x}_i = \bm{o} + s_i \bm{d}$ has a distance $s_i$ from the camera center and a distance of $\delta_i = s_{i+1} - s_i$ between its adjacent sample $\bm{x}_{i+1}$.

\subsection{Event Generation Model}
\label{subsec:method:egm}

\begin{figure}[t]
    \begin{center}
    \includegraphics[width=0.65\linewidth,trim={0.4cm 0.2cm 0.4cm 0.2cm},clip]{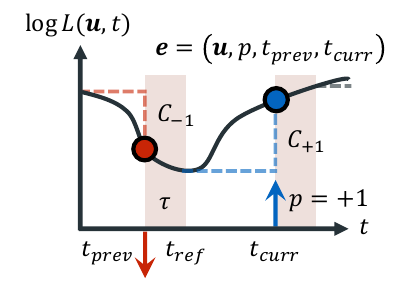}
    \end{center}
    \vspace{-0.75em}
    \caption{Event generation model. An event $\bm{e}$ of polarity $p$ is generated at timestamp $t_{\mathit{curr}}$ when the difference in log-radiance $\log L$ at a pixel $\bm{u}$, measured with respect to a reference $\log L$ at timestamp $t_{\mathit{ref}}$, has the same sign as $p$ and a magnitude that equals to the contrast threshold associated to polarity $p$, $C_p$. Red, downwards and blue, upwards arrows represent events of polarities $-1$ and $+1$, respectively, and each right-angled dashed line represents the measured change in $\log L$. After an event is generated, the pixel will be temporarily deactivated for an amount of time given by the refractory period $\tau$, as shaded in the figure. Thus, $t_{\mathit{ref}}$ is simply the sum of the previous event timestamp $t_{\mathit{prev}}$ and $\tau$.}
    \vspace{-0.65em}
    \label{fig:egm}
\end{figure}

\begin{figure*}[t]
    \begin{center}
    \includegraphics[width=1\linewidth,trim={0.4cm 0.5cm 0.0cm 0.2cm},clip]{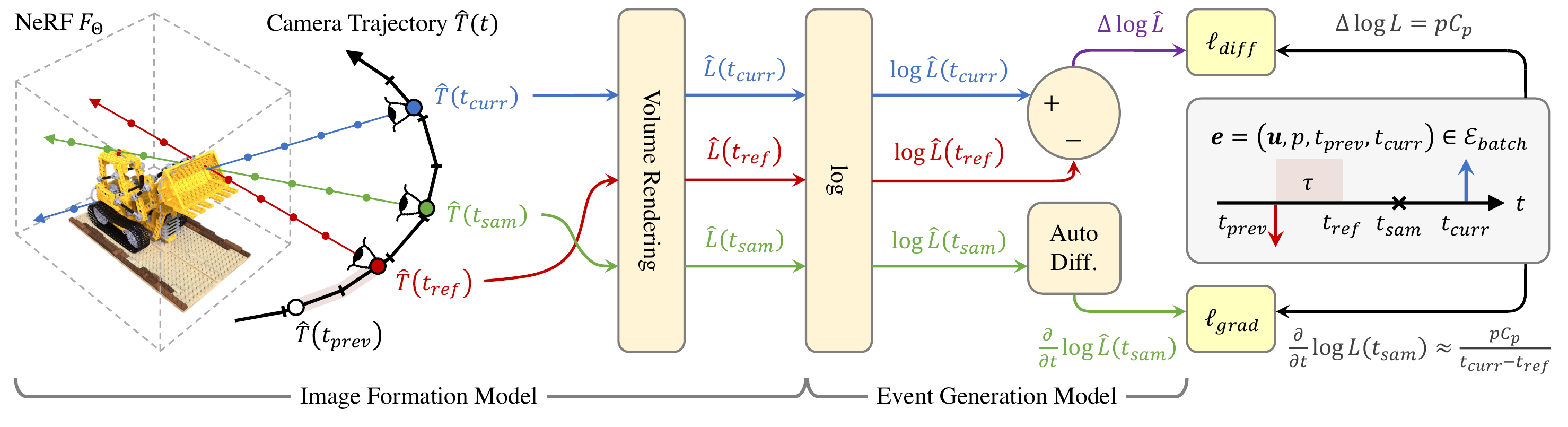}
    \end{center}
    % \vspace{-0.75em}
    \caption{Overview of Robust \textit{e}-NeRF training. For each event $\bm{e}$ in the batch $\mathcal{E}_{\mathit{batch}}$ sampled randomly from the event stream, we first derive the reference timestamp $t_{\mathit{ref}}$ (\autoref{eq:t_ref}), given the refractory period $\tau$, and sample a timestamp $t_{\mathit{sam}}$ between $t_{\mathit{ref}}$ and $t_{\mathit{curr}}$. Next, we interpolate the given constant-rate camera poses at $t_{\mathit{ref}}, t_{\mathit{sam}}$ and $t_{\mathit{curr}}$ using LERP for position and SLERP for orientation. Given these pose estimates $\hat{T}$, we then perform volume rendering on the back-projected rays from pixel $\bm{u}$ with the NeRF $F_\Theta$ (\autoref{eq:L-hat}). This is done to infer the predicted radiance $\hat{L}$, and thus log-radiance $\log{\hat{L}}$, of pixel $\bm{u}$ at $t_{\mathit{ref}}, t_{\mathit{sam}}$ and $t_{\mathit{curr}}$. For brevity, we denote $\hat{L} (t) = \hat{L} (\bm{u}, t)$. These $\log{\hat{L}}$ are ultimately used to derive the predicted log-radiance difference $\Delta \log \hat{L}$ and gradient $\frac{\partial}{\partial t} \log \hat{L} (t_{\mathit{sam}})$ for the computation of the reconstruction loss: threshold-normalized difference loss $\ell_{\mathit{diff}}$ 
(\autoref{eq:l_diff}) and smoothness loss: target-normalized gradient loss $\ell_{\mathit{grad}}$ (\autoref{eq:l_grad}), given the observed log-radiance difference $\Delta \log L$ (\autoref{eq:delta_logL}) and gradient $\frac{\partial}{\partial t} \log L (t_{\mathit{sam}})$ approximation from the event $\bm{e}$, respectively.}

% as the image formation model, event generation model
   
    \vspace{-0.65em}
    \label{fig:overview}
\end{figure*}

\autoref{fig:egm}
shows the illustration of the event generation model.
An event camera responds to log-radiance changes in the scene and outputs an \textit{Event Stream} $\mathcal{E}$, given by:
\begin{equation}
	\mathcal{E} = \{ \ \bm{e} \mid \bm{e} = (\bm{u}, p, t_{\mathit{prev}}, t_{\mathit{curr}}) \ \} \ ,
    \label{eq:event_stream}
\end{equation}
where $\bm{e}$ is an \textit{Event} generated by pixel $\bm{u}$, with polarity $p \in \{ -1, +1 \}$, at timestamp $t_{\mathit{curr}}$. For convenience of discussion, we augment each event with the timestamp of the previous event that was generated by the same pixel, $t_{\mathit{prev}}$.

An event of polarity $p$ is generated when the difference in log-radiance at a pixel, measured with respect to a reference log-radiance at timestamp $t_{\mathit{ref}}$, has the same sign as $p$ and a magnitude that equals to the \textit{Contrast Threshold} associated to polarity $p$, $C_p$ \cite{gallego2020_event_survey}. In short, the condition is given by:
\begin{equation}
	\Delta \log L \defeq \log L (\bm{u}, t_{\mathit{curr}}) - \log L (\bm{u}, t_{\mathit{ref}}) = pC_p \ ,
    \label{eq:delta_logL}
\end{equation}
where $L$ denotes the incident radiance at the given pixel and timestamp. For color event cameras, $L$ corresponds to the radiance of the incident light after passing through the specific color filter in front of the pixel.

After an event is generated, the pixel will be deactivated for an amount of time specified by the \textit{Refractory Period} $\tau$. During this period of time, the pixel is invariant to any change in log-radiance and thus will not generate any new events. At the end of the refractory period, the pixel will be reactivated and the current log-radiance value at the pixel will be set as the new reference value, enabling the next event to be generated at this pixel \cite{gallego2020_event_survey,lichtsteiner2008_128}. In essence: 
\begin{equation}
    t_{\mathit{ref}} = t_{\mathit{prev}} + \tau \ .
    \label{eq:t_ref}
\end{equation}

The refractory period gives rise to temporal sparsity in the event stream. While this leads to partially observable log-radiance changes, it also limits the event generation rate of the camera \cite{gallego2020_event_survey, lichtsteiner2008_128}. This is crucial to prevent the corruption of event timestamps due to \textit{Address Event Representation} (AER) bus saturation and readout congestion, especially when the resolution and/or the speed of the camera are high \cite{gallego2020_event_survey, inivation2020_whitepaper, posch2014_retinomorphic, finateu2020_prophesee}. The bounded event rate also facilitates longer periods of recordings given a fixed memory budget. Furthermore, long refractory periods, as well as asymmetric contrast thresholds, contribute to a lower \textit{Shot Noise} event rate \cite{mcreynolds2023_exploiting}, thereby improving \textit{Signal-to-Noise Ratio} (SNR).

Event sensors also suffer from non-uniformity of pixel response, similar to standard image sensors. This is characterized by the pixel-to-pixel variation in the contrast threshold, which can be viewed as the event sensor analogue of \textit{Fixed-Pattern Noise} (FPN). Studies suggest that the threshold variation can be modeled as a \textit{Gaussian} distribution with mean $C_p$ \cite{gallego2020_event_survey, lichtsteiner2008_128, posch2011_sensitivity}.

The contrast threshold is also effectively constant over time, as it is independent of temperature. The key assumption for temperature-independence is the absence of junction leakage \cite{nozaki2017_temperature}, which holds true for modern event cameras \cite{son2017_samsung}. Regardless of leakage, temperature-independence remains observed over a large temperature range \cite{nozaki2017_temperature}.

\subsection{Training}
\label{subsec:method:training}

\subsubsection{Assumptions}
\label{subsubsec:method:assmp}

To reconstruct a NeRF from an event camera in motion, we assume that we are given the intrinsic camera matrix, lens distortion parameters and constant-rate camera poses of sufficiently high sampling rate for accurate interpolation at arbitrary time instants. Similar to \cite{klenk2022_enerf}, we perform \textit{Linear Interpolation} (LERP) on camera positions and \textit{Spherical Linear Interpolation} (SLERP) on camera orientations. While our method does not rely on event camera-specific intrinsic parameters such as contrast threshold and refractory period to be known as \textit{a priori}, %their knowledge will 
this information generally facilitates a more accurate reconstruction. As we will see in \autoref{subsubsec:method:limitations}, it is also acceptable to provide only the \textit{(positive-to-negative) contrast threshold ratio} $\nicefrac{C_{+1}}{C_{-1}}$, which can be inferred from event camera bias settings.

\vspace{-0.65em}
\subsubsection{Fundamental Limitations}
\label{subsubsec:method:limitations}

As event cameras only provide (partial) observations of log-radiance \emph{changes}, not \emph{absolute} log-radiance, the predicted log-radiance $\log \hat{\bm{L}}$, given by volume renderings from the reconstructed NeRF (\autoref{eq:L-hat}), is only accurate up to an offset per color channel. Furthermore, there will be an additional scale ambiguity that is consistent across all channels, when only the contrast threshold ratio is known. This is similar to an image with unknown \textit{black levels} and ISO. Nonetheless, these ambiguities can be easily dealt with or corrected for post-reconstruction, \eg using a set of reference images of the same scene (\autoref{subsec:method:gamma}), thereby not a cause for concern. We are also restricted to the reconstruction of a NeRF with single-channel directional emitted radiance, when a monochrome event camera is employed.

\vspace{-0.65em}
\subsubsection{Loss Functions}
\label{subsubsec:method:losses}

An overview of the training pipeline is illustrated in \autoref{fig:overview}.
In line with the event generation model (\autoref{subsec:method:egm}), we propose two complementary normalized loss functions: \textit{threshold-normalized difference loss} $\ell_{\mathit{diff}}$ and \textit{target-normalized gradient loss} $\ell_{\mathit{grad}}$ that directly and effectively generalize to various real-world conditions. The weighted sum of the two losses form the total training loss, which is optimized on a batch of events $\mathcal{E}_{\mathit{batch}}$ sampled randomly from the raw, asynchronous event stream. %In short
Formally, the total training loss is given by:
\begin{equation}
	\mathcal{L} = \frac{1}{\abs{\mathcal{E}_{\mathit{batch}}}} \sum_{\bm{e} \in \mathcal{E}_{\mathit{batch}}} \lambda_{\mathit{diff}} \ell_{\mathit{diff}} (\bm{e}) + \lambda_{\mathit{grad}} \ell_{\mathit{grad}} (\bm{e}) \ ,
    \label{eq:total_loss}
\end{equation}
where $\lambda_{\mathit{diff}}$ and $\lambda_{\mathit{grad}}$ are the respective loss weights.

We specifically refrained from optimizing on a reduced event stream, obtained by accumulating successive events at each pixel over time intervals, as done in \cite{rudnev2022_eventnerf, hwang2022_evnerf, klenk2022_enerf}. This enables us to account for the refractory period and prevent unnecessary effective reduction in contrast sensitivity, which leads to lower reconstruction fidelity. Moreover, it can also be shown that event accumulation results in the effective amplification of threshold variation (proof in supplementary materials), thereby reduction in noise robustness.

\vspace{-0.65em}
\paragraph{Threshold-Normalized Difference Loss.}

This loss enforces the \textit{mean contrast threshold} $\bar{C} = \frac{1}{2} (C_{-1} + C_{+1})$ normalized squared consistency between the observed log-radiance difference $\Delta \log L = p C_p$ from an event (\autoref{eq:delta_logL}) and the predicted log-radiance difference $\Delta \log \hat{L} \defeq \log \hat{L} (\bm{u}, t_{\mathit{curr}}) - \log \hat{L} (\bm{u}, t_{\mathit{ref}})$, given by renders from the NeRF model (\autoref{eq:L-hat}), as follows:
\begin{equation}
	\ell_{\mathit{diff}} (\bm{e}) = \left( \frac{\Delta \log \hat{L} - p C_p}{\bar{C}} \right)^2 \ .
    \label{eq:l_diff}
\end{equation}
Note that when a color event camera is employed, $\hat{L}$ refers to the single-channel rendered radiance, where 
the color channel is governed by the pixel color filter.

The loss serves as the primary reconstruction loss and can be effectively interpreted as a squared percentage error, especially under symmetric contrast thresholds. The normalization entails that the loss is invariant to the common scale of the positive and negative thresholds, as well as the predicted log-radiance, and only dependent on their ratio. Moreover, the normalization is optimal in the sense that the magnitude of the normalized target $\abs*{\nicefrac{p C_p}{\bar{C}}} = \nicefrac{C_p}{\bar{C}}$ is always centered at 1 regardless of the threshold ratio (proof in supplementary materials). %Thus, 
The loss %is able to 
can therefore effectively generalize to arbitrary threshold values.

Unlike Ev-NeRF \cite{hwang2022_evnerf}, these properties also enable the joint optimization of the unknown contrast threshold without additional regularization, as it does not suffer from any degeneracy. However, only the contrast threshold ratio can be recovered, thereby an additional scale ambiguity in the predicted log radiance, as mentioned in \autoref{subsubsec:method:limitations}. In addition, our experiments also demonstrate the viability of jointly optimizing the refractory period $\tau$ via $t_\mathit{ref}$ (\autoref{eq:t_ref}).

\vspace{-0.65em}
\paragraph{Target-Normalized Gradient Loss.}

This loss is simply the \textit{Absolute Percentage Error} (APE) of the predicted log-radiance temporal gradient derived using auto-differentiation $\frac{\partial}{\partial t} \log \hat{L} (\bm{u}, t)$, with respect to the finite difference approximation of the target log-radiance gradient $\frac{\partial}{\partial t} \log L (\bm{u}, t) \approx \frac{p C_p}{t_{\mathit{curr}} - t_{\mathit{ref}}}$, at a timestamp $t_{\mathit{sam}}$ sampled between $t_{\mathit{ref}}$ and $t_{\mathit{curr}}$, as follows:
\begin{equation}
	\ell_{\mathit{grad}} (\bm{e}) = \scalebox{0.95}{$\operatorname{APE}$} \left(\frac{\partial}{\partial t} \log \hat{L} (\bm{u}, t_{\mathit{sam}}) \fraccomma \ \frac{p C_p}{t_{\mathit{curr}} - t_{\mathit{ref}}} \right)
    \label{eq:l_grad}
\end{equation}
where $\operatorname{APE} \left(\hat{y}, y \right) = \abs*{\frac{\hat{y} - y}{y}}$. As the finite difference approximation error is minimum at the midpoint and maximum at the endpoints, we sample $t_{\mathit{sam}}$ from a truncated normal distribution that is centered at the midpoint and has a standard deviation of $\nicefrac{1}{4}$ the interval.

The loss acts as a smoothness constraint for log-radiance changes between $t_{\mathit{ref}}$ and $t_{\mathit{curr}}$, which lack explicit regularization from $\ell_{\mathit{diff}}$. This is particularly important for the effective reconstruction of textureless regions in the scene, where events associated have comparably longer intervals. In contrast to analogous regularization losses adopted in recent works \cite{rudnev2022_eventnerf, hwang2022_evnerf, klenk2022_enerf}, $\ell_{\mathit{grad}}$ is specifically invariant to the speed of motion, hence generalizable to arbitrary speed profiles. An unnormalized gradient loss would over-emphasize events generated under high-speed motion, as they have relatively larger target gradients. %Apart from that
Furthermore, the loss is invariant to the common scale of the threshold and predicted log-radiance, similar to $\ell_{\mathit{diff}}$. %Thus, 
It is therefore also able to effectively generalize to arbitrary threshold values and facilitate joint optimization of unknown threshold.

\subsection{Gamma Correction of Synthesized Views}
\label{subsec:method:gamma}

As alluded in \autoref{subsubsec:method:limitations}, the channel-consistent scale and per-channel offset ambiguity in the predicted log-radiance can be corrected for post-reconstruction, given a set of reference images of the same scene. To better account for the likely mismatch of \textit{spectral sensitivity}, thereby \textit{color balance}, between the event camera and the standard camera used to capture the reference images. Akin to \cite{rudnev2022_eventnerf}, we further relax the channel-consistent scale to per-channel scales. This entails %that 
an \textit{affine} correction of the predicted \emph{log}-radiance, or equivalently a \textit{gamma} correction of the predicted \emph{linear} radiance, for each color channel as follows:
\begin{equation}
    \log \hat{\bm{L}}_\mathit{corr} = \bm{a} \odot \log \hat{\bm{L}} + \bm{b} \ ,
    \label{eq:log_L-hat_corr}
\end{equation}
where $\bm{a}$ and $\bm{b}$  are the correction parameters, is necessary and sufficient for the reference alignment. %Hence
Consequently, given the target log-radiance $\log \bm{L}$ from the reference images, the optimal correction parameters can be simply derived via \textit{ordinary least squares}.

\section{Experiments}
\label{sec:exp}

We adopt \textit{Novel View Synthesis} (NVS) as the standard task to verify that our method, Robust \textit{e}-NeRF can indeed directly and robustly reconstruct NeRFs from moving event cameras under various real-world conditions, particularly from sparse and noisy events generated under non-uniform motion. The NVS benchmark experiments are conducted on both synthetic (\autoref{subsec:exp:synthetic}) and real sequences (\autoref{subsec:exp:real}). In addition, we perform ablation studies (\autoref{subsec:exp:ablation}) to investigate the significance of various components in our method.

\vspace{-0.65em}
\paragraph{Metrics.}

We adopt a consistent set of metrics to assess the performance of a method in all experiments. In particular, we employ the three following standard metrics: \textit{Peak Signal-to-Noise Ratio} (PSNR), \textit{Structural Similarity Index Measure} (SSIM) \cite{zhou2004_ssim} and AlexNet-based \textit{Learned Perceptual Image Patch Similarity} (LPIPS) \cite{zhang2018_lpips} to quantify the similarity between the gamma-corrected (\autoref{subsec:method:gamma}) synthesized novel views and given target novel views.

\vspace{-0.65em}
\paragraph{Baselines.}

We benchmark our method, Robust \textit{e}-NeRF against a recent work, Ev-NeRF \cite{hwang2022_evnerf} and a naïve baseline, E2VID $+$ NeRF, given by cascading the seminal events-to-video reconstruction method, E2VID \cite{rebecq2021_e2vid} to NeRF. For Ev-NeRF, events are accumulated over non-overlapping time intervals of $\nicefrac{1}{24} \ s \approx 41.67 \ \mathit{ms}$, as suggested in their paper. To facilitate a fair comparison, all methods, including ours, have been (re-)implemented to adopt a common NeRF backbone. While we do not explicitly compare against E-NeRF \cite{klenk2022_enerf} and EventNeRF \cite{rudnev2022_eventnerf}, our experiment results should still provide an accurate indication on their performance relative to ours since they share a lot of similarities with Ev-NeRF as detailed in \autoref{sec:intro}.

\vspace{-0.65em}
\paragraph{Datasets.}

The synthetic experiments are performed on a novel set of sequences simulated on the ``Realistic Synthetic $360^\circ$'' scenes, which were adopted in the synthetic experiments of NeRF \cite{mildenhall2020_nerf}. These scenes contain a wide variety of photo-realistic objects with complicated structure and non-Lambertian effects, thereby effective for NVS evaluation. The new synthetic event dataset allows for a retrospective comparison between event-based and image-based NeRF reconstruction methods, as the sequences were simulated under highly similar conditions, unlike in \cite{rudnev2022_eventnerf, hwang2022_evnerf}.

% same test views

The events are generated from a virtual event camera moving in a hemi-/spherical spiral motion about the object at the origin, as shown in \autoref{fig:teaser}(b, d), using an improved version of ESIM \cite{rebecq2018_esim}, which is an efficient and realistic event simulator. Specifically, we added support to time-independent pixel-to-pixel threshold variation and Blender, circumvented singularities in the trajectory orientation interpolation, as well as improved event simulation accuracy, especially with non-zero refractory periods. On the contrary, the real experiments are performed on the \texttt{mocap-1d-trans} and \texttt{mocap-desk2} sequences of the TUM-VIE dataset \cite{klenk2021_tumvie}, which are mainly forward-facing captures of a desk with various objects on top, under linear and spiral camera motion, respectively. These real sequences were chosen for their relative suitability for NVS and their adoption of a modern, high-resolution event sensor --- Prophesee Gen 4.

\subsection{Synthetic Experiments}
\label{subsec:exp:synthetic}

The synthetic experiments form the core of the benchmark, as they allow for realistic \emph{controlled} experiments under various real-world conditions with \emph{absolute} ground truth, which are otherwise infeasible using real sequences.

All sequences are simulated with a symmetric contrast threshold of $0.25$ (\ie $C_{-1} = C_{+1} = 0.25$), which is the approximate nominal threshold for event sensors such as the Prophesee Gen 3.1, Gen 4.1 and Sony IMX636 sensor, hence providing a known threshold ratio of 1 (\ie $\nicefrac{C_{+1}}{C_{-1}} = 1$). The event camera trajectory is also sampled at a high rate of $1 \ \mathit{kHz}$ to minimize the influence of pose errors on the performance assessment. Furthermore, the virtual event camera revolves the object 4 times, with uniform 1 revolution per second speed about the object vertical axis by default, similar to \cite{rudnev2022_eventnerf}. Unless otherwise stated, a sequence is also simulated with zero pixel-to-pixel threshold standard deviation and refractory period (\ie $\sigma_{C_p} = 0, \tau = 0$). By default, Ev-NeRF and our method are trained with a constant symmetric contrast threshold, given by the prior knowledge of time-independent threshold ratio. 

\begin{table*}[t!]
\small
\setlength{\tabcolsep}{2.8pt}

\begin{center}
\begin{tabular}{lccclccclccclccc}
\toprule
\multicolumn{1}{c}{} & \multicolumn{3}{c}{$v = 1 \times$} &  & \multicolumn{3}{c}{$v_b = 8 \times$} &  & \multicolumn{3}{c}{$v = \frac{1}{8} \times$} &  & \multicolumn{3}{c}{$v = 8 \times$} \\ \cmidrule(lr){2-4} \cmidrule(lr){6-8} \cmidrule(lr){10-12} \cmidrule(l){14-16} 
\multicolumn{1}{c}{\multirow{-2}{*}{Method}} & PSNR $\uparrow$ & SSIM $\uparrow$ & LPIPS $\downarrow$ &  & PSNR $\uparrow$ & SSIM $\uparrow$ & LPIPS $\downarrow$ &  & PSNR $\uparrow$ & SSIM $\uparrow$ & LPIPS $\downarrow$ &  & PSNR $\uparrow$ & SSIM $\uparrow$ & LPIPS $\downarrow$ \\ \midrule
E2VID $+$ NeRF & 18.92 & 0.832 & 0.316 &  & 18.92 & 0.832 & 0.316 &  & 18.92 & 0.832 & 0.316 &  & 18.92 & 0.832 & 0.316 \\
Ev-NeRF & 27.72 & 0.935 & 0.087 &  & 26.25 & 0.926 & 0.102 &  & 19.79 & 0.792 & 0.326 &  & 20.83 & 0.862 & 0.198 \\
\rowcolor[HTML]{F3F3F3} 
Robust \textit{e}-NeRF & \textbf{28.19} & \textbf{0.945} & \textbf{0.057} &  & \textbf{28.19} & \textbf{0.945} & \textbf{0.057} &  & \textbf{28.19} & \textbf{0.945} & \textbf{0.057} &  & \textbf{28.19} & \textbf{0.945} & \textbf{0.057} \\ \bottomrule
\end{tabular}
\end{center}

\caption{Effect of speed profile. $v$ denotes the speed of motion relative to the default hemi-/spherical spiral motion with uniform azimuth speed, whereas $v_b$ denotes the oscillation factor of the relative speed of motion (\ie $v = {v_b}^{\sin{2 \pi f t} }, f = 1 \mathit{Hz}$).}
\label{tab:speed}
\end{table*}

\begin{table*}[t!]
\small
\setlength{\tabcolsep}{6.2pt}

\begin{center}
\begin{tabular}{lcccclccclccc}
\toprule
\multicolumn{1}{c}{} &  & \multicolumn{3}{c}{$\sigma_{C_p} = 0.00$} &  & \multicolumn{3}{c}{$\sigma_{C_p} = 0.03$} &  & \multicolumn{3}{c}{$\sigma_{C_p} = 0.06$} \\ \cmidrule(lr){3-5} \cmidrule(lr){7-9} \cmidrule(l){11-13} 
\multicolumn{1}{c}{\multirow{-2}{*}{Method}} & \multirow{-2}{*}{\begin{tabular}[c]{@{}c@{}}Opt.\\ $C_p$\end{tabular}} & PSNR $\uparrow$ & SSIM $\uparrow$ & LPIPS $\downarrow$ &  & PSNR $\uparrow$ & SSIM $\uparrow$ & LPIPS $\downarrow$ &  & PSNR $\uparrow$ & SSIM $\uparrow$ & LPIPS $\downarrow$ \\ \midrule
E2VID $+$ NeRF & $-$ & 18.92 & 0.832 & 0.316 &  & 18.68 & 0.827 & 0.330 &  & 18.03 & 0.808 & 0.363 \\
 & $\times$ & 27.72 & 0.935 & 0.087 &  & 24.42 & 0.895 & 0.155 &  & 8.07 & 0.841 & 0.260 \\
\multirow{-2}{*}{Ev-NeRF} & $\checkmark$ & 27.43 & 0.911 & 0.123 &  & 23.66 & 0.826 & 0.261 &  & 15.43 & 0.708 & 0.441 \\
\rowcolor[HTML]{F3F3F3} 
\cellcolor[HTML]{F3F3F3} & $\times$ & \textbf{28.19} & \textbf{0.945} & \textbf{0.057} &  & \textbf{28.14} & \textbf{0.946} & \textbf{0.058} &  & \textbf{28.23} & \textbf{0.947} & \textbf{0.057} \\
\rowcolor[HTML]{F3F3F3} 
\multirow{-2}{*}{\cellcolor[HTML]{F3F3F3}Robust \textit{e}-NeRF} & $\checkmark$ & \textbf{28.17} & \textbf{0.946} & \textbf{0.051} &  & \textbf{27.91} & \textbf{0.946} & \textbf{0.054} &  & \textbf{28.19} & \textbf{0.948} & \textbf{0.049} \\ \bottomrule
\end{tabular}
\end{center}

\caption{Effect of pixel-to-pixel threshold variation $\sigma_{C_p}$. ``Opt. $C_p$'' refers to jointly optimizing thresholds $C_p$ with NeRF parameters $\Theta$.}
\label{tab:threshold_var}
\end{table*}

\begin{table*}[t!]
\small
\setlength{\tabcolsep}{5.2pt}

\begin{center}
\begin{tabular}{lccccclccclccc}
\toprule
\multicolumn{1}{c}{} &  &  & \multicolumn{3}{c}{$\tau = 0 \mathit{ms}$} &  & \multicolumn{3}{c}{$\tau = 8 \mathit{ms}$} &  & \multicolumn{3}{c}{$\tau = 25 \mathit{ms}$} \\ \cmidrule(lr){4-6} \cmidrule(lr){8-10} \cmidrule(l){12-14} 
\multicolumn{1}{c}{\multirow{-2}{*}{Method}} & \multirow{-2}{*}{\begin{tabular}[c]{@{}c@{}}Opt.\\ $C_p$\end{tabular}} & \multirow{-2}{*}{\begin{tabular}[c]{@{}c@{}}Opt.\\ $\tau$\end{tabular}} & PSNR $\uparrow$ & SSIM $\uparrow$ & LPIPS $\downarrow$ &  & PSNR $\uparrow$ & SSIM $\uparrow$ & LPIPS $\downarrow$ &  & PSNR $\uparrow$ & SSIM $\uparrow$ & LPIPS $\downarrow$ \\ \midrule
E2VID $+$ NeRF & $-$ & $-$ & 18.92 & 0.832 & 0.316 &  & 14.87 & 0.797 & 0.427 &  & 14.15 & 0.791 & 0.467 \\
 & $\times$ & $-$ & 27.72 & 0.935 & 0.087 &  & 13.17 & 0.707 & 0.559 &  & 12.75 & 0.759 & 0.528 \\
\multirow{-2}{*}{Ev-NeRF} & $\checkmark$ & $-$ & 27.43 & 0.911 & 0.123 &  & 13.56 & 0.716 & 0.528 &  & 13.75 & 0.717 & 0.569 \\
\rowcolor[HTML]{F3F3F3} 
\cellcolor[HTML]{F3F3F3} & $\times$ & $\times$ & \textbf{28.19} & \textbf{0.945} & \textbf{0.057} &  & \textbf{26.30} & \textbf{0.934} & \textbf{0.066} &  & \textbf{25.51} & \textbf{0.929} & \textbf{0.072} \\
\rowcolor[HTML]{F3F3F3} 
\multirow{-2}{*}{\cellcolor[HTML]{F3F3F3}Robust \textit{e}-NeRF} & $\times$ & $\checkmark$ & \textbf{28.18} & \textbf{0.945} & \textbf{0.052} &  & \textbf{23.43} & \textbf{0.910} & \textbf{0.090} &  & \textbf{22.48} & \textbf{0.895} & \textbf{0.105} \\ \bottomrule
\end{tabular}
\end{center}

\caption{Effect of refractory period $\tau$. ``Opt. $\tau$'' refers to jointly optimizing refractory period $\tau$ with NeRF parameters $\Theta$.}
\label{tab:refr}
\end{table*}

\begin{table*}[t!]
\small
\setlength{\tabcolsep}{4.6pt}

\begin{center}
\begin{tabular}{lccccclccclccc}
\toprule
\multicolumn{1}{c}{} &  &  & \multicolumn{3}{c}{\footnotesize $v_b = 1 \times, \sigma_{C_p} = 0.00, \tau = 0 \mathit{ms}$} &  & \multicolumn{3}{c}{\footnotesize $v_b = 4 \times, \sigma_{C_p} = 0.03, \tau = 8 \mathit{ms}$} &  & \multicolumn{3}{c}{\footnotesize $v_b = 8 \times, \sigma_{C_p} = 0.06, \tau = 25 \mathit{ms}$} \\ \cmidrule(lr){4-6} \cmidrule(lr){8-10} \cmidrule(l){12-14} 
\multicolumn{1}{c}{\multirow{-2}{*}{Method}} & \multirow{-2}{*}{\begin{tabular}[c]{@{}c@{}}Opt.\\ $C_p$\end{tabular}} & \multirow{-2}{*}{\begin{tabular}[c]{@{}c@{}}Opt.\\ $\tau$\end{tabular}} & PSNR $\uparrow$ & SSIM $\uparrow$ & LPIPS $\downarrow$ &  & PSNR $\uparrow$ & SSIM $\uparrow$ & LPIPS $\downarrow$ &  & PSNR $\uparrow$ & SSIM $\uparrow$ & LPIPS $\downarrow$ \\ \midrule
E2VID $+$ NeRF & $-$ & $-$ & 18.92 & 0.832 & 0.316 &  & 14.98 & 0.796 & 0.433 &  & 14.07 & 0.801 & 0.448 \\
 & $\times$ & $-$ & 27.72 & 0.935 & 0.087 &  & 12.33 & 0.742 & 0.521 &  & 12.05 & 0.807 & 0.425 \\
\multirow{-2}{*}{Ev-NeRF} & $\checkmark$ & $-$ & 27.43 & 0.911 & 0.123 &  & 13.06 & 0.732 & 0.539 &  & 12.27 & 0.772 & 0.539 \\
\rowcolor[HTML]{F3F3F3} 
\cellcolor[HTML]{F3F3F3} & $\times$ & $\times$ & \textbf{28.19} & \textbf{0.945} & \textbf{0.057} &  & \textbf{24.10} & \textbf{0.913} & \textbf{0.086} &  & \textbf{23.51} & \textbf{0.900} & \textbf{0.110} \\
\rowcolor[HTML]{F3F3F3} 
\multirow{-2}{*}{\cellcolor[HTML]{F3F3F3}Robust \textit{e}-NeRF} & $\checkmark$ & $\checkmark$ & \textbf{28.19} & \textbf{0.946} & \textbf{0.051} &  & \textbf{20.42} & \textbf{0.875} & \textbf{0.126} &  & \textbf{18.83} & \textbf{0.836} & \textbf{0.197} \\ \bottomrule
\end{tabular}
\end{center}

\caption{Collective effect of speed profile, threshold variation and refractory period.}
\label{tab:collective}
\end{table*}

\begin{table*}[t!]
\small
\setlength{\tabcolsep}{8pt}

\begin{center}
\begin{tabular}{ccccclccclccc}
\toprule
 &  & \multicolumn{3}{c}{\small $v_b = 1 \times, \sigma_{C_p} = 0.00, \tau = 0 \mathit{ms}$} &  & \multicolumn{3}{c}{\small $v_b = 4 \times, \sigma_{C_p} = 0.03, \tau = 8 \mathit{ms}$} &  & \multicolumn{3}{c}{\small $v_b = 8 \times, \sigma_{C_p} = 0.06, \tau = 25 \mathit{ms}$} \\ \cmidrule(lr){3-5} \cmidrule(lr){7-9} \cmidrule(l){11-13} 
\multirow{-2}{*}{$\tau$} & \multirow{-2}{*}{$\ell_\mathit{grad}$} & PSNR $\uparrow$ & SSIM $\uparrow$ & LPIPS $\downarrow$ &  & PSNR $\uparrow$ & SSIM $\uparrow$ & LPIPS $\downarrow$ &  & PSNR $\uparrow$ & SSIM $\uparrow$ & LPIPS $\downarrow$ \\ \midrule
$\times$ & $\checkmark$ & \textbf{28.19} & \textbf{0.945} & \textbf{0.057} &  & 12.77 & 0.799 & 0.372 &  & 12.41 & 0.798 & 0.412 \\
$\checkmark$ & $\times$ & 27.96 & 0.943 & 0.063 &  & 23.15 & 0.899 & 0.113 &  & 22.21 & 0.879 & 0.153 \\
\rowcolor[HTML]{F3F3F3} 
$\checkmark$ & $\checkmark$ & \textbf{28.19} & \textbf{0.945} & \textbf{0.057} &  & \textbf{24.10} & \textbf{0.913} & \textbf{0.086} &  & \textbf{23.51} & \textbf{0.900} & \textbf{0.110} \\ \bottomrule
\end{tabular}
\end{center}

\caption{Ablation studies on refractory period $\tau$ modeling and target-normalized gradient loss $\ell_\mathit{grad}$.}

\label{tab:ablation}
\end{table*}

\begin{figure*}[t!]
    \begin{center}
    \includegraphics[width=1\linewidth]{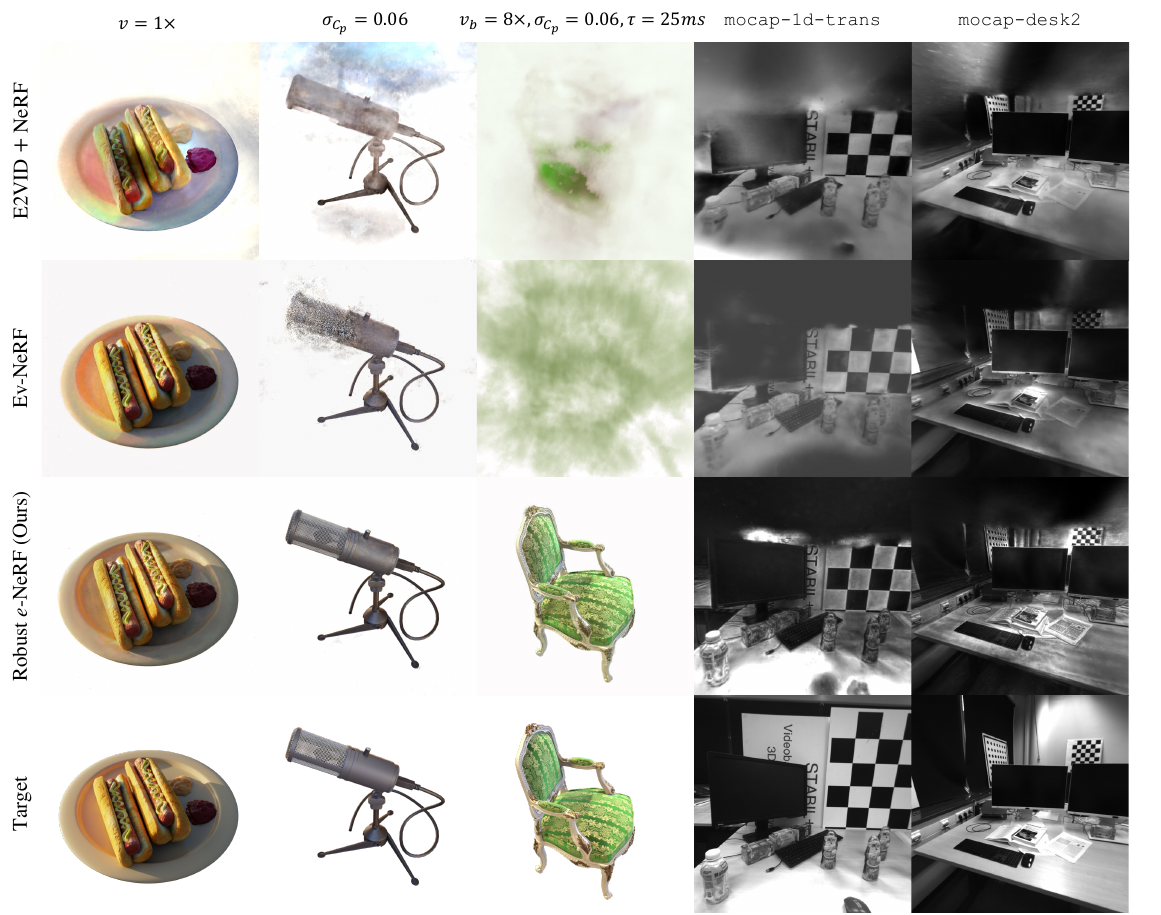}
    \end{center}

    \caption{Synthesized novel views in the synthetic and real experiments. Results shown correspond to that with jointly optimized contrast thresholds and possibly refractory period, if applicable for the method in the particular experiment.}
    \vspace{-0.65em}
    \label{fig:qualitative}
\end{figure*}

\vspace{-0.65em}
\paragraph{Effect of Speed Profile.}

To investigate the influence of camera speed, we evaluate all works on 4 sets of sequences simulated with different speed profiles, where each set contains a sequence for each of the 7 scenes. Particularly, the 1\textsuperscript{st} set is simulated with the default settings, thus providing a reference (uniform azimuth) speed of motion (\ie $v = 1\times$). We oscillate the speed of motion between $\nicefrac{1}{8} \times$ and $8 \times$ the original speed at a frequency of $1 \mathit{Hz}$ (\ie $v = {v_b}^{\sin{2 \pi f t} }$, where $v_b = 8 \times$ and $f = 1 \mathit{Hz}$) for the 2\textsuperscript{nd} set, and scale the speed of motion to $\nicefrac{1}{8} \times$ and $8 \times$ of the original speed (\ie $v = \nicefrac{1}{8} \times, 8 \times$) for the 3\textsuperscript{rd} and 4\textsuperscript{th} sets, respectively. \autoref{fig:teaser}(b) illustrates the uniform motion in the 1\textsuperscript{st}, 3\textsuperscript{rd} and 4\textsuperscript{th} sets, whereas \autoref{fig:teaser}(d) illustrates the non-uniform, oscillating motion in the 2\textsuperscript{nd} set.

The quantitative results reported in \autoref{tab:speed} underscores the significance of speed invariance in enabling the effective generalization to arbitrary speed profiles, as it can be observed that the performance of Ev-NeRF deteriorates as the speed deviates from the optimal at $v = 1 \times$. Our method also outperforms all baselines, including Ev-NeRF, at the default setting (\ie $v = 1 \times$), which is optimal for all. Qualitative results shown in \autoref{fig:qualitative} suggests an overall improvement in synthesis quality, including high-frequency details.

\vspace{-0.65em}
\paragraph{Effect of Pixel-to-Pixel Threshold Variation.}

To evaluate the robustness of our method to noise, in the form of threshold variation, we benchmark all works on three sets of sequences. Furthermore, we also benchmark Ev-NeRF and our method with jointly optimized contrast thresholds, which are poorly initialized with $\nicefrac{C_{+1}}{C_{-1}} = 10$ (more precisely, we set $C_{-1} = 0.25$ and $C_{+1} = 2.5$), to assess the robustness of the joint optimization. The three sets of sequences were simulated with pixel-to-pixel threshold standard deviations of 0, 0.03 and 0.06 (\ie $\sigma_{C_p} = 0.00, 0.03, 0.06$), which are equivalent to $0 \%$, $12 \%$ and $24 \%$ of the contrast threshold, respectively. $\sigma_{C_p} = 0.03$ is approximately the nominal value for the Prophesee Gen 3.1, Gen 4 and Sony IMX636 sensor, and $\sigma_{C_p} = 0.06$ is the maximum value for the Sony IMX636 sensor. 

The quantitative results given in \autoref{tab:threshold_var} clearly demonstrates the robustness of our method to threshold variation, as our performance is essentially unaffected by the degree of noise. In contrast, the baselines are visibly susceptible to threshold variation, as their performance declines with its severity, especially Ev-NeRF. It is interesting to note that even the naïve baseline outperforms Ev-NeRF at $\sigma_{C_p} = 0.06$. These conclusions are also supported qualitatively in \autoref{fig:qualitative}. The quantitative results also illustrate the effectiveness and robustness of our threshold joint optimization, as the performance of our method with and without jointly optimized thresholds are virtually the same. In contrast, it can be observed that the joint optimization of time-varying contrast thresholds in Ev-NeRF leads to a slight decrease in performance on the default setting (\ie $\sigma_{C_p} = 0.00$).

\vspace{-0.65em}
\paragraph{Effect of Refractory Period.}

To study the effect of temporal sparsity in the event stream due to the refractory period, we benchmark all methods on three sets of sequences simulated with refractory periods of $0$, $8$ and $25 
 \mathit{ms}$ (\ie $\tau = 0, 8, 25 \mathit{ms}$ ). Similar to the previous experiment, we additionally evaluate our method with jointly optimized refractory period, which is poorly initialized with half the maximum possible value, given by the minimum time interval between successive events at a pixel in the sequence. Furthermore, we also benchmark Ev-NeRF with jointly optimized thresholds to validate the importance of accounting for the refractory period. $\tau = 8 \mathit{ms}$ is just slightly less than mean event interval in the set of sequences with $\tau = 0 \mathit{ms}$. Let $\mathcal{E}_\tau$ be the event stream with refractory period $\tau$, the degree of sparsity due to $\tau$ can be quantified by the mean $\nicefrac{\abs{\mathcal{E}_{\tau = 0 \mathit{ms}}}}{\abs{\mathcal{E}_\tau}}$ across all sequences in the set, which translates to $5.84 \times$ and $11.37 \times$ for $\tau = 8$ and $25 \mathit{ms}$, respectively.

The quantitative results presented in \autoref{tab:refr} undoubtedly verifies the importance of refractory period modeling in enabling the robust reconstruction from temporally sparse event streams, as our method significantly outperforms all baselines across all values of $\tau$. Moreover, the attempt by Ev-NeRF to compensate $\tau$ with jointly optimized thresholds has also clearly failed, albeit contributing to slightly improved performance. It is also worth noting that the naïve baseline achieves better reconstructions than Ev-NeRF under non-zero $\tau$. Our perceivable drop in performance, as $\tau$ increases, may be attributed to its associated decrease in observability. Our results also demonstrate the feasibility of jointly optimizing $\tau$, although less effective than threshold joint optimization, which reflects the complexities involved.

\vspace{-0.65em}
\paragraph{Collective Effect.}

To assess the performance of all methods under the collective effect of threshold variation, refractory period and camera speed, which resembles real-world operating conditions, we benchmark all works on three sets of sequences with different levels of difficulty: \textit{easy} ($\sigma_{C_p} = 0.00, \tau = 0 \mathit{ms}, v_b = 1 \times$), \textit{medium} ($\sigma_{C_p} = 0.03, \tau = 8 \mathit{ms}, v_b = 4 \times$) and \textit{hard} ($\sigma_{C_p} = 0.06, \tau = 25 \mathit{ms}, v_b = 8 \times$). In addition, we also benchmark Ev-NeRF and our method with jointly optimized intrinsic parameters, similar to previous experiments. The medium and hard sets are $5.31 \times$ and $8.27 \times$ sparser than the easy set, respectively. This experiment is particularly challenging as a $K\times$ speed-up effectively scales $\tau$ by $K \times$ in the context of the original sequence. Moreover, methods that involve event accumulation will also experience a $K\times$ effective scaling in $\sigma_{C_p}$.

The quantitative results provided in \autoref{tab:collective} generally reflect that of the previous experiment, thus the same conclusions can be drawn. This should not be surprising, given the proven importance of accounting for $\tau$. Our decline in performance on the medium and hard settings, when compared to $\tau = 8$ and $25 \mathit{ms}$, may be attributed to the complications arising from the interaction between $\sigma_{C_p}$, $\tau$ and $v_b$. Qualitative results given in \autoref{fig:qualitative} illustrates the strength and robustness of our method under challenging conditions.

\subsection{Real Experiments}
\label{subsec:exp:real}

The real experiments mainly serve as a qualitative benchmark, as the dataset (and other similar ones) is not specifically suited for the task of NVS, thereby unable to provide an accurate quantification of performance (refer to supplementary materials for a detailed justification). For the real experiments, Ev-NeRF and our method are trained with jointly optimized intrinsic parameters, where the contrast threshold is initialized with $\nicefrac{C_{+1}}{C_{-1}} = 1$ (more precisely, we set $C_{-1} = C_{+1} = 0.25$) this time. While the two sequences mainly involve an insignificant refractory period relative to their primarily non-uniform speed of motion, qualitative results shown in \autoref{fig:qualitative} still demonstrate our superior performance in recovering fine details while appropriately smoothing uniform regions, in line with the synthetic experiments. Exposure of the scene is also visibly more consistent and more accurately reconstructed. Visible artifacts near the borders of all synthesized views are due to the comparably narrower \textit{field-of-view} of the event camera.

\subsection{Ablation Studies}
\label{subsec:exp:ablation}

The ablation studies are conducted in the same setting as the synthetic experiment on the collective effect, without jointly optimized intrinsic parameters. The quantitative results reported in \autoref{tab:ablation} further validates the immense importance of accounting for $\tau$, even when event accumulation is not involved. Although the synthetic sequences mainly involves scenes with abundant texture, the results also verify the effectiveness of the target-normalized gradient loss in regularizing textureless regions, particularly under more challenging conditions (\ie medium and hard settings).

\section{Conclusion}
\label{sec:conclusion}

In this paper, we introduce Robust \textit{e}-NeRF, a novel approach to directly and robustly reconstruct NeRFs from moving event cameras under various real-world conditions, particularly from sparse and noisy events generated under non-uniform motion. Our method consists of two key components: a realistic event generation model that accounts for various event camera-specific intrinsic parameters and non-idealities, as well as a complementary pair of normalized reconstruction losses that can effectively generalize to arbitrary speed profiles and intrinsic parameter values without such prior knowledge. The proposed \textit{Analysis-by-Synthesis} approach for event-based reconstruction naturally extends to other scene representations, such as 3D Gaussians \cite{kerbl2023_3d_gaussian_splatting}. In spite of its achievements, Robust \textit{e}-NeRF still relies on the assumption of a static scene and given constant-rate camera poses, which we leave for future work. We also see the modeling of other spatial/temporal noise and artifacts of an event sensor as a promising future direction for more accurate reconstructions.

\vspace{-0.65em}
\paragraph{Acknowledgements.} This research is supported by the National Research Foundation, Singapore under its AI Singapore Programme (AISG Award No: AISG2-RP-2021-024),
and the Tier 2 grant MOE-T2EP20120-0011 from the Singapore Ministry of Education.

{\small
\bibliographystyle{ieee_fullname}
\bibliography{references}
}

% ---- Supplementary Materials ----
\clearpage

\title{Supplementary Material for\\Robust \textit{e}-NeRF: NeRF from Sparse \& Noisy Events under Non-Uniform Motion}

\author{Weng Fei Low \qquad Gim Hee Lee\\
The NUS Graduate School's Integrative Sciences and Engineering Programme (ISEP)\\
Institute of Data Science (IDS), National University of Singapore\\
Department of Computer Science, National University of Singapore\\
{\tt\small \{wengfei.low, gimhee.lee\}@comp.nus.edu.sg}\\
\small{\url{https://wengflow.github.io/robust-e-nerf}}
% \\
}

\maketitle
\ificcvfinal\thispagestyle{empty}\fi

\appendix
\setcounter{equation}{8}
\setcounter{table}{5}
\setcounter{figure}{4}

%%%%%%%%% BODY TEXT
In this supplementary document, we first show how event accumulation results in an effective amplification of pixel-to-pixel contrast threshold variation (\autoref{sec:amplif_var}) and discuss the optimality of normalization in the threshold-normalized difference loss (\autoref{sec:l_diff_norm_optimality}). Next, we present implementation details of Robust \textit{e}-NeRF and the baselines used in the experiments (\autoref{sec:impl_details}). We also provide a detailed justification on the qualitative nature of our real experiments (\autoref{sec:justif_real}). Lastly, we present additional quantitative and qualitative results on all experiments (\autoref{sec:add_results}).

\section{Amplification of Threshold Variation}
\label{sec:amplif_var}

As alluded in Sec.~3.3.3, the accumulation of successive events at each pixel over time intervals leads to the effective amplification of pixel-to-pixel contrast threshold variation. This can be shown by simply analyzing the distribution of the target log-radiance difference after event accumulation, at any given pixel.

The time-independent contrast threshold of polarity $p$ can be modeled as a random variable $c_p \sim \mathcal{N}(C_p, {\sigma_{C_p}}^2)$ (Sec.~3.2). Assuming $N_p$ number of polarity $p$ events are accumulated at the pixel within the specified time interval, the target log-radiance difference $\Delta \log L_\mathit{acc}$ is then given by:
\begin{equation}
    \Delta \log L_\mathit{acc} = \sum_p p N_p c_p \ ,
    \label{eq:delta_log_L_acc}
\end{equation}
which follows the Gaussian distribution below:
\begin{equation}
    \mathcal{N} \left( \sum_p p N_p C_p, \ \sum_p {N_p}^2 {\sigma_{C_p}}^2 - 2 N_{+1} N_{-1} \sigma_{c_{+1}, c_{-1}} \right) \ ,
    \label{eq:delta_log_L_acc_dist}
\end{equation}
where $\sigma_{c_{+1}, c_{-1}} \in \left[ -\sigma_{c_{+1}} \sigma_{c_{-1}}, \ \sigma_{c_{+1}} \sigma_{c_{-1}} \right]$ is the covariance between $c_{+1}$ and $c_{-1}$.

Note that when $N_{+1}$ and $N_{-1}$ increases by a factor of $K$, the standard deviation of $\Delta \log L_\mathit{acc}$ will also increase by the same factor, which results in noisier targets. Moreover, assuming that  $c_{+1}$ and $c_{-1}$ do not have a strong positive correlation (\ie  $\sigma_{c_{+1}, c_{-1}} \ll \sigma_{c_{+1}} \sigma_{c_{-1}}$, with respect to the range of $\sigma_{c_{+1} c_{-1}}$), which is highly likely to be true, it can also be shown that standard deviation of $\Delta \log L_\mathit{acc} \gg \abs{\sum_p p N_p \sigma_{C_p}} \ge 0$ under non-zero $N_{+1}$ and $N_{-1}$. This suggests that when $N_{+1} C_{+1} \approx N_{-1} C_{-1}$, which often holds true in practice over sufficiently long accumulation intervals (relative to the speed of motion and amount of scene texture), the mean of $\Delta \log L_\mathit{acc} = \sum_p p N_p C_p \approx 0$ whereas the standard deviation remains very much larger than 0, especially for large $N_{+1}$ and $N_{-1}$. Such a cancellation between positive and negative accumulated events further aggravates the target noise. All these observations suggest an effective amplification of threshold variation when event accumulation is involved.

\section{Optimality of Normalization in $\ell_\mathit{diff}$}
\label{sec:l_diff_norm_optimality}

As mentioned in Sec.~3.3.3, the threshold-normalized difference loss $\ell_\mathit{diff}$ (Eq.~6) is optimal in the sense that the magnitude of the normalized target $\abs*{\nicefrac{p C_p}{\bar{C}}}$, which is essentially the normalized threshold $\nicefrac{C_p}{\bar{C}}$, is always centered at 1 regardless of the threshold ratio $\nicefrac{C_{+1}}{C_{-1}}$, as follows:
\begin{equation}
    \abs*{\frac{p C_p}{\bar{C}}} = \frac{C_p}{\bar{C}} = 1 + p \frac{\tilde{C}}{\bar{C}}
    \label{eq:l_diff_norm_optimality}
\end{equation}
where $\tilde{C} = \frac{1}{2} (C_{+1} - C_{-1})$ and the magnitude of the offset $\nicefrac{\tilde{C}}{\bar{C}}$ can be interpreted as the normalized threshold difference. This facilitates the scale consistency of the loss, thus enabling the adoption of a single, global loss weight $\lambda_\mathit{diff}$ for arbitrary contrast threshold values. Nevertheless, the variance of the normalized target increases as the thresholds become more asymmetric.

\section{Implementation Details}
\label{sec:impl_details}

\subsection{Robust \textit{e}-NeRF}
\label{subsec:impl_details:ours}

\paragraph{Architecture.}

Robust \textit{e}-NeRF adopts Instant-NGP \cite{muller2022_instant_ngp} as the NeRF backbone, as it allows for high-quality reconstructions given relatively low training time and memory cost. More precisely, we employ the implementation provided by the NerfAcc toolbox \cite{li2022_nerfacc}, due to its simple and flexible Python APIs, but with some slight modifications.

In particular, parameters of the embedded \textit{Multi-Layer Perceptron} (MLP) are initialized using the PyTorch-default method, instead of \textit{Xavier} initialization \cite{glorot2010_xavier}. Furthermore, we replace all \textit{Rectified Linear Unit} (ReLU) hidden layer activations with \textit{SoftPlus} ($\beta = 100$) as it is infinitely differentiable everywhere, thereby facilitating the optimization of $\ell_\mathit{grad}$.

Since the predicted \emph{log}-radiance is at most accurate up to an offset per color channel (Sec.~3.3.2), or equivalently the predicted \emph{linear} radiance (modeled by NeRF) is at most accurate up to a scale per color channel, we also replace the bounded sigmoid radiance output activation with the lower-bounded SoftPlus (default $\beta = 1$). In addition, we add a small $\epsilon = 0.001$ to the positive raw radiance output from the NeRF model (\ie $\hat{\bm{L}} = \hat{\bm{L}}_\mathit{raw} + \epsilon$) to improve the numerical stability of the predicted log-radiance $\log \hat{\bm{L}}$. This augmentation imposes a lower bound of $\epsilon$ on the radiance our method can \emph{model}, as $\hat{\bm{L}} > \epsilon$. Nevertheless, this is not a cause for concern given the minimum per-channel scale ambiguity of $\hat{\bm{L}}$, non-upper bounded range of $\hat{\bm{L}}_\mathit{raw}$ and non-zero scene radiance (\ie absolute darkness is virtually impossible in practice).

For synthetic scenes, we also alpha composite $\hat{\bm{L}}_\mathit{raw}$ with a learnable background radiance, which is parameterized via SoftPlus to ensure that it is always positive, prior to $\epsilon$-augmentation. In contrast, common NeRF backbones and EventNeRF \cite{rudnev2022_eventnerf} adopt a fixed background, which is inappropriate given the scale ambiguity.

As only the threshold ratio can be recovered during the joint optimization of contrast threshold (Sec.~3.3.3), we keep the negative threshold $C_{-1}$ fixed at an arbitrary value and only optimize the learnable positive-to-negative contrast threshold ratio $\nicefrac{C_{+1}}{C_{-1}}$, which is parameterized via SoftPlus to ensure that it is always positive. Moreover, since the refractory period is lower bounded at 0 and upper bounded by the minumum time interval between successive events at any pixel (Sec.~4.1), we parameterize the refractory period via a scaled sigmoid that preserves the gradient profile of the default, unscaled sigmoid function. We additionally clamp the parameterized refractory period between $\varepsilon$ and $(1 - \varepsilon) \times$ its range to limit the minimum gradient of the scaled sigmoid to approximately $\varepsilon \times$ the range. This prevents vanishing gradients at the extremes, which implicates the optimization of the refractory period.

For real scenes, we appropriately predefine the \textit{Axis-Aligned Bounding Box} (AABB), as well as the near and far bounds of the back-projected rays used for volume rendering, for each scene. Furthermore, we employ the spherical space contraction proposed in mip-NeRF 360 \cite{barron2022_mipnerf360} to better model unbounded scenes. We also increase the occupancy grid resolution to $256^3$ and set the cone angle (\ie ray marching step size increment scale) to $0.004$, which is approximately $\nicefrac{1}{256}$ as suggested by Instant-NGP.

\paragraph{Training.}

The training loss weights used in all experiments are given by $\lambda_\mathit{diff} = 1$ and $\lambda_\mathit{grad} = 0.001$. As suggested by Instant-NGP, we also impose a weight decay of $10^{-6}$ on the MLP to prevent overfitting. The model is trained for 40 000 iterations with a learning rate decay of 0.33 at 20 000, 30 000 and 36 000 iterations (\ie 50\%, 75\% and 90\% progress, as done in NerfAcc), using the Adam optimizer \cite{kingma2015_adam} with a learning rate of 0.01 and PyTorch-default hyper-parameters. During joint optimization of contrast threshold, its parameter is assigned a higher learning rate of 0.1 to facilitate to its early convergence. Moreover, since the scaled sigmoid function preserves its gradient profile, but the range of the refractory period may vary greatly, the learning rate assigned to the (unscaled logit) parameter of refractory period is set to $50 \times$ the range. The event batch size is determined dynamically based on the average number of ray samples used to render a single pixel, similar to Instant-NGP, to maximize the utilization of the GPU memory. Specifically, we ensure that every batch of events involves approximately $2^{20} = 1\ 048\ 576$ samples in total, for either the rays at $t_\mathit{ref}$, $t_\mathit{curr}$ (relevant to $\ell_\mathit{diff}$) \emph{or} $t_\mathit{sam}$ (relevant to $\ell_\mathit{grad}$). As a side note, the poses of the target novel views in the real experiments are interpolated from the given unsynchronized constant-rate camera poses using LERP and SLERP.

\subsection{Baselines}
\label{subsec:impl_details:baselines}

As alluded in Sec.~4, both baselines have been carefully reimplemented on the same NerfAcc backbone and trained with the same hyper-parameters (including the weight decay), when applicable, to facilitate a fair comparison. However, we only train the naïve baseline of E2VID $+$ NeRF for 20 000 iterations with a learning rate decay of 0.33 at 10 000, 15 000 and 18 000 iterations (\ie 50\%, 75\% and 90\% progress) due to its comparably faster convergence, as a result of the direct absolute radiance supervision. Similar to the target novel views, the poses of the E2VID-reconstructed training views are also interpolated from the given unsynchronized constant rate camera poses using LERP and SLERP. Furthermore, we extend the implementation of E2VID to support the RGGB \textit{Bayer} pattern adopted in ESIM.

\section{Justification of Qualitative Real Exps.}
\label{sec:justif_real}

As mentioned in Sec.~4.2, we mainly perform qualitative evaluation for the real experiments. This is done because the target novel views, given by a separate standard camera, suffer from saturation due to the comparably smaller dynamic range of the standard camera, and are not raw images that have not been processed by the lossy in-camera image processing pipeline. Moreover, the spectral sensitivity curve of the event camera adopted is also not documented, hence gamma correction may not accurately align the synthesized views.

Furthermore, the comparably narrower \textit{field-of-view} of the event camera and the limited camera motion also leads to a relatively smaller coverage of the scene, thereby causing artifacts in the synthesized novel views near the borders, as observed in the qualitative results. This further complicates the quantitative evaluation as it is non-trivial to delineate the valid synthesis regions. Other event camera datasets also suffer from similar issues, as all are not specifically suited for novel view synthesis.

\section{Additional Experiment Results}
\label{sec:add_results}

\subsection{Per-Scene Breakdown}
\label{sec:add_results:breakdown}

\begin{figure*}[t!]
    \begin{center}
    \includegraphics[width=1\linewidth]{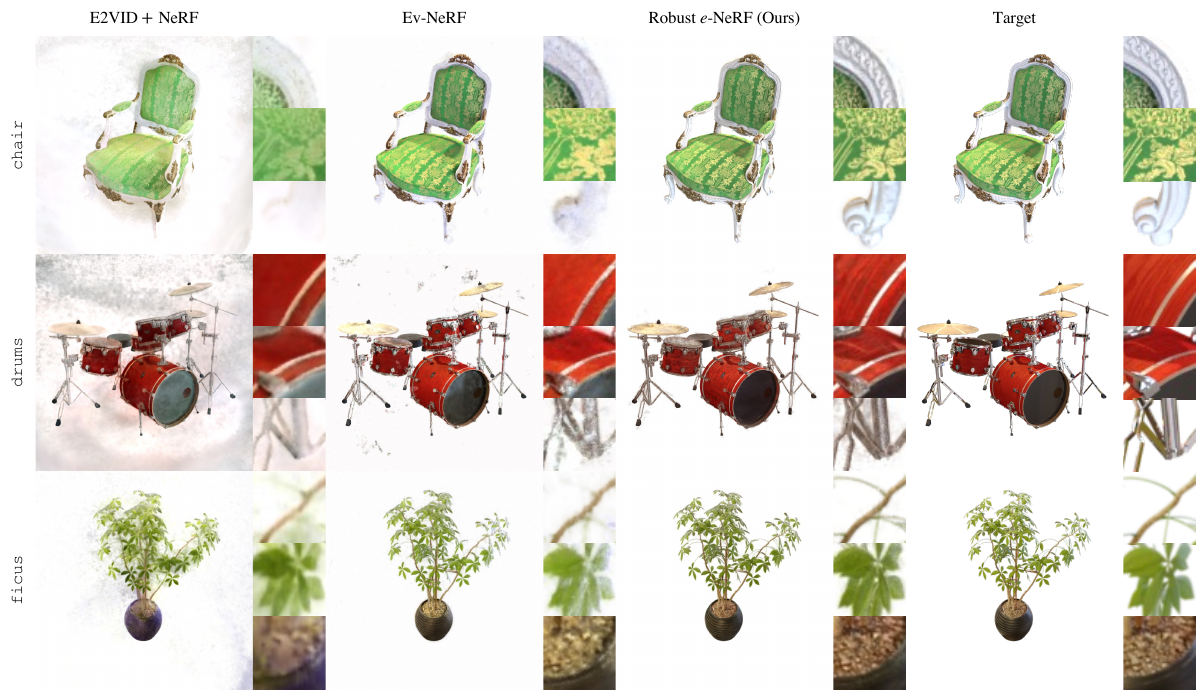}
    \end{center}

    \caption{Synthesized novel views on \texttt{chair},  \texttt{drums} and \texttt{ficus} under the default setting.}
    \label{fig:breakdown1}
\end{figure*}

\begin{figure*}[t!]
    \begin{center}
    \includegraphics[width=1\linewidth]{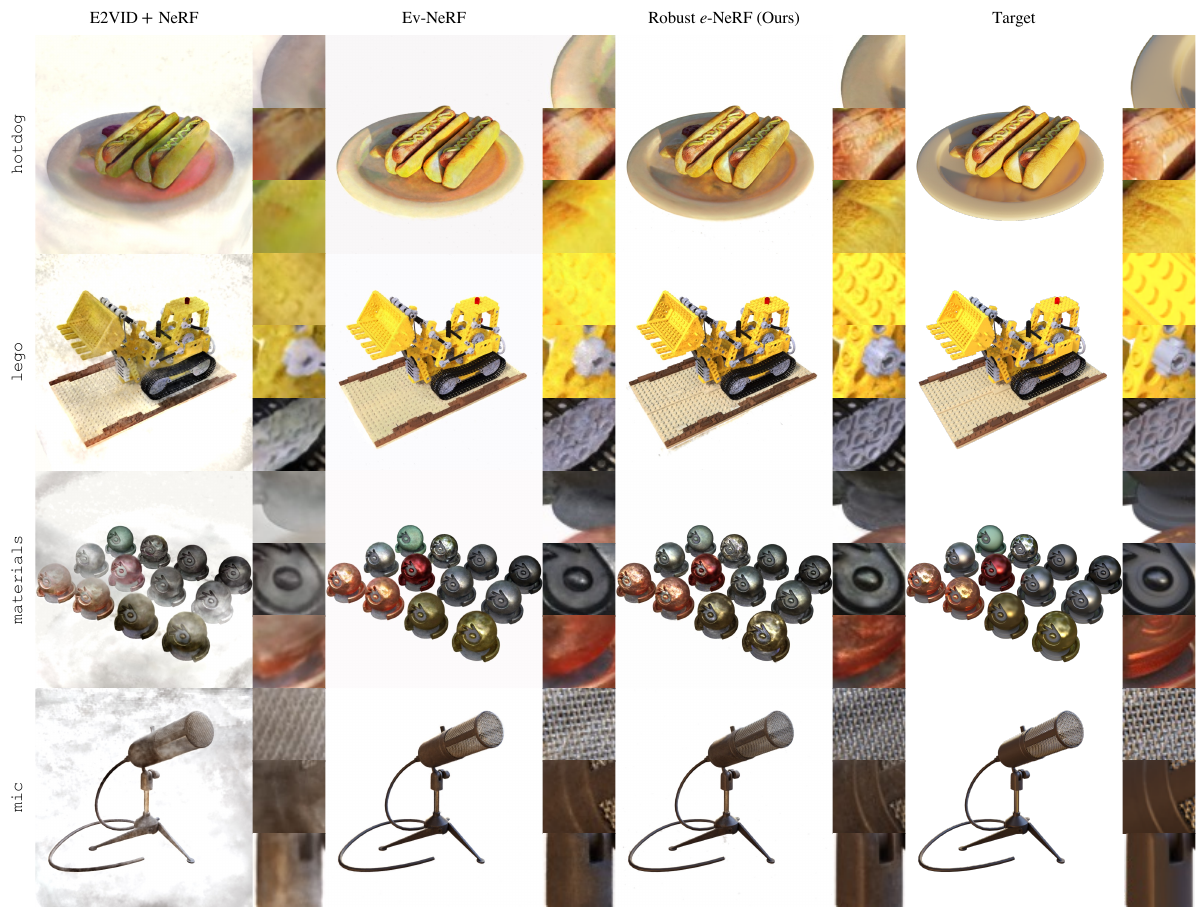}
    \end{center}

    \caption{Synthesized novel views on \texttt{hotdog}, \texttt{lego},  \texttt{materials} and \texttt{mic} under the default setting.}
    \label{fig:breakdown2}
\end{figure*}

\autoref{tab:breakdown} and \autoref{fig:breakdown1}, \ref{fig:breakdown2} show the quantitative and qualitative results of all methods, respectively, for each of the seven synthetic scene sequences simulated with the default settings, which is optimal for all methods. The per-scene quantitative results is generally consistent with the aggregate metrics, which is also presented in Sec.~4.1, as our method outperforms the baselines in most scenes and has comparable performance in others. The per-scene qualitative results reveal our superior performance in reconstructing fine details and maintaining high color accuracy, especially at the background, as previously observed in Sec.~4.1.

\begin{table*}[t!]
% \small
\setlength{\tabcolsep}{7.5pt}

\begin{center}
\begin{tabular}{llcccccccc}
\toprule
\multicolumn{1}{c}{} & \multicolumn{1}{c}{} & \multicolumn{7}{c}{Synthetic Scene} &  \\ \cmidrule(lr){3-9}
\multicolumn{1}{c}{\multirow{-2}{*}{Metric}} & \multicolumn{1}{c}{\multirow{-2}{*}{Method}} & \texttt{chair} & \texttt{drums} & \texttt{ficus} & \texttt{hotdog} & \texttt{lego} & \texttt{materials} & \texttt{mic} & \multirow{-2}{*}{Mean} \\ \midrule
 & E2VID $+$ NeRF & 19.62 & 19.52 & 22.44 & 17.33 & 17.41 & 18.13 & 18.02 & 18.92 \\
 & Ev-NeRF & 28.93 & \textbf{23.89} & 28.37 & 25.22 & \textbf{29.10} & \textbf{26.50} & 32.03 & 27.72 \\
\multirow{-3}{*}{PSNR $\uparrow$} & \cellcolor[HTML]{F3F3F3}Robust \textit{e}-NeRF & \cellcolor[HTML]{F3F3F3}\textbf{30.24} & \cellcolor[HTML]{F3F3F3}23.15 & \cellcolor[HTML]{F3F3F3}\textbf{30.71} & \cellcolor[HTML]{F3F3F3}\textbf{28.07} & \cellcolor[HTML]{F3F3F3}27.34 & \cellcolor[HTML]{F3F3F3}24.98 & \cellcolor[HTML]{F3F3F3}\textbf{32.87} & \cellcolor[HTML]{F3F3F3}\textbf{28.19} \\ \midrule
 & E2VID $+$ NeRF & 0.869 & 0.842 & 0.863 & 0.859 & 0.710 & 0.835 & 0.844 & 0.832 \\
 & Ev-NeRF & 0.932 & 0.889 & 0.948 & 0.940 & 0.930 & \textbf{0.926} & 0.979 & 0.935 \\
\multirow{-3}{*}{SSIM $\uparrow$} & \cellcolor[HTML]{F3F3F3}Robust \textit{e}-NeRF & \cellcolor[HTML]{F3F3F3}\textbf{0.958} & \cellcolor[HTML]{F3F3F3}\textbf{0.897} & \cellcolor[HTML]{F3F3F3}\textbf{0.971} & \cellcolor[HTML]{F3F3F3}\textbf{0.953} & \cellcolor[HTML]{F3F3F3}\textbf{0.934} & \cellcolor[HTML]{F3F3F3}0.923 & \cellcolor[HTML]{F3F3F3}\textbf{0.981} & \cellcolor[HTML]{F3F3F3}\textbf{0.945} \\ \midrule
 & E2VID $+$ NeRF & 0.277 & 0.277 & 0.289 & 0.341 & 0.406 & 0.282 & 0.337 & 0.316 \\
 & Ev-NeRF & 0.085 & 0.203 & 0.085 & 0.103 & \textbf{0.058} & 0.054 & \textbf{0.024} & 0.087 \\
\multirow{-3}{*}{LPIPS $\downarrow$} & \cellcolor[HTML]{F3F3F3}Robust \textit{e}-NeRF & \cellcolor[HTML]{F3F3F3}\textbf{0.040} & \cellcolor[HTML]{F3F3F3}\textbf{0.091} & \cellcolor[HTML]{F3F3F3}\textbf{0.022} & \cellcolor[HTML]{F3F3F3}\textbf{0.095} & \cellcolor[HTML]{F3F3F3}0.074 & \cellcolor[HTML]{F3F3F3}\textbf{0.052} & \cellcolor[HTML]{F3F3F3}0.029 & \cellcolor[HTML]{F3F3F3}\textbf{0.057} \\ \bottomrule
\end{tabular}
\end{center}

\caption{Per-synthetic scene breakdown under the default setting.}
\label{tab:breakdown}
\end{table*}

\subsection{Qualitative Analysis of $\ell_\mathit{grad}$}
\label{sec:add_results:qual_l_grad}

\begin{figure*}[t!]
    \begin{center}
    \includegraphics[width=1\linewidth]{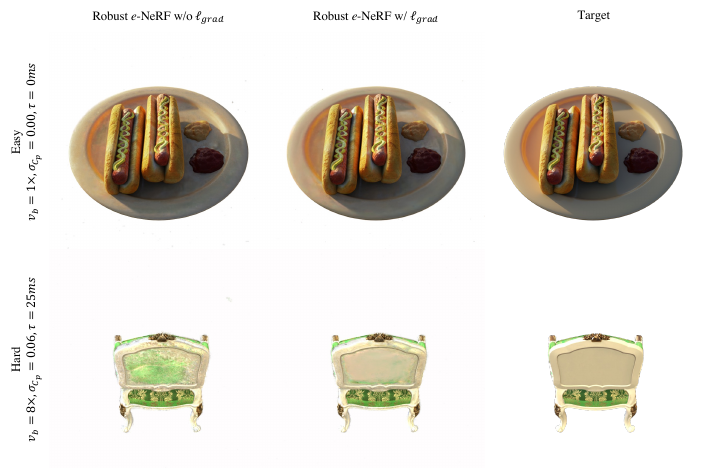}
    \end{center}

    \caption{Synthesized novel views with and without the target-normalized gradient loss $\ell_\mathit{grad}$.}
    \label{fig:qual_l_grad}
\end{figure*}

\autoref{fig:qual_l_grad} illustrates the effect of target-normalized gradient loss $\ell_\mathit{grad}$ on the \texttt{hotdog} and \texttt{chair} scene sequences simulated with the easy and hard settings, respectively, as similarly done in Sec.~4.3. It can be observed that with $\ell_\mathit{grad}$, the plate of the hotdog and the back of the chair exhibit less noise, especially the latter. This is achieved while preserving high-frequency details on the hotdog and the cushion of the chair. This further validates the effectiveness of $\ell_\mathit{grad}$ in regularizing textureless regions, particularly under challenging conditions.

\subsection{Qualitative Results on \texttt{office-maze}}
\label{sec:add_results:office-maze}

\begin{figure*}[t!]
    \begin{center}
    \includegraphics[width=1\linewidth]{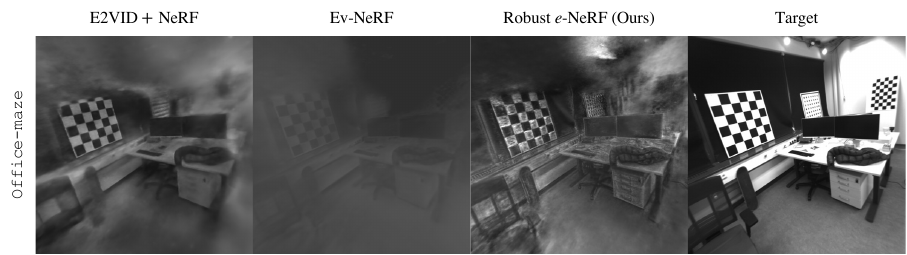}
    \end{center}

    \caption{Synthesized novel views on the \texttt{office-maze} scene.}
    \label{fig:office-maze}
\end{figure*}

Apart from \texttt{mocap-1d-trans} and \texttt{mocap-desk2}, we also benchmark all methods on the \texttt{office-maze} sequence from the TUM-VIE dataset. We only employ the subsequence before the 395\textsuperscript{th} target novel view, as it captures a bounded space of an office (in approximately 2 loops around the office). The qualitative results reported in \autoref{fig:office-maze} clearly shows our effectiveness in recovering details and resolving the scene structure without suffering from severe fogs in free space.

\subsection{Robustness to Temporal Event Sparsity}
\label{sec:add_results:efficiency}

To evaluate the robustness of our method to temporal sparsity of the event stream (\ie data efficiency), we benchmark it on a set of nine sequences simulated on the synthetic \texttt{chair} scene with different refractory periods. Apart from the standard image similarity performance metrics, we also report some statistics such as the percentage of $\tau$ relative to the duration of the event sequence, as well as the degree of sparsity of the event stream, as defined in Sec.~4.1. Moreover, we also report the number of images that occupy an equivalent amount of memory as the event sequence disregarding compression, assuming 8 bits per image pixel channel and 47 bits per event (\ie $2 \times 11$ bits for position, $1$ bit for polarity and $24$ bits for timestamp), as implied after decompression of the Prophesee EVT 3.0 \cite{evt3} event encoding format.

\begin{figure*}[t!]
    \begin{center}
    \includegraphics[width=0.42\linewidth]{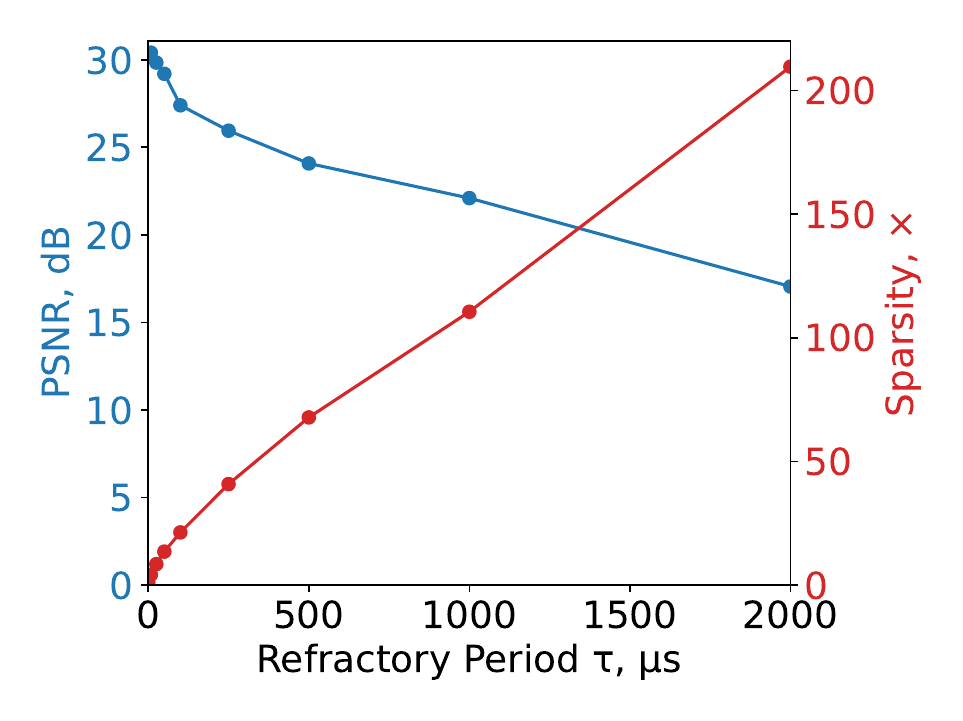}
    \end{center}

    \caption{Plot of novel view synthesis PSNR and degree of event sparsity on the \texttt{chair} scene against refractory period $\tau$.}
    \label{fig:efficiency_plot}
\end{figure*}

\begin{figure*}[t!]
    \begin{center}
    \includegraphics[width=1\linewidth]{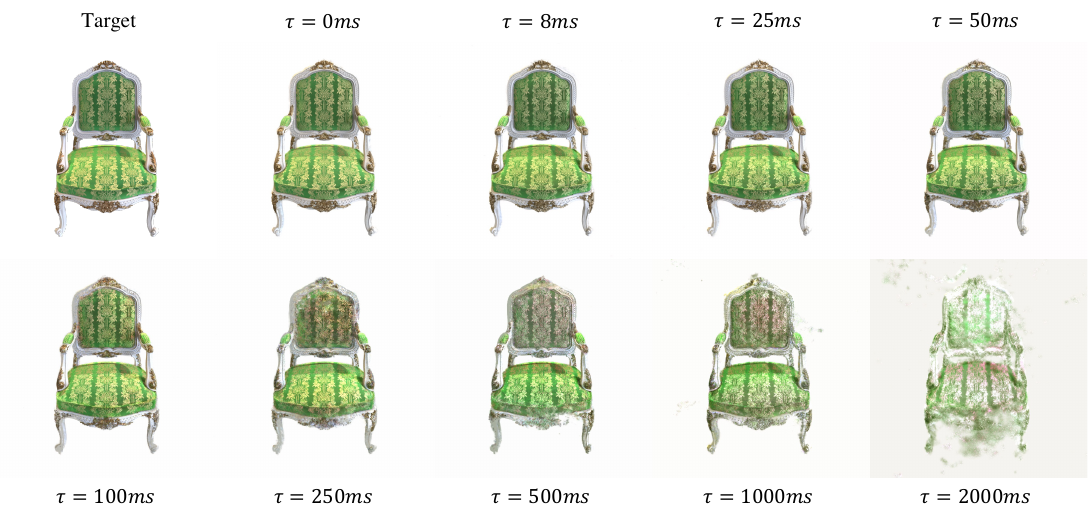}
    \end{center}

    \caption{Synthesized novel views on the \texttt{chair} scene under numerous refractory periods $\tau$.}
    \label{fig:efficiency}
\end{figure*}

The quantitative and qualitative results given in \autoref{tab:efficiency}, \autoref{fig:efficiency_plot} and \autoref{fig:efficiency} demonstrate our astonishing robustness under severely sparse event streams, which suggests that our method is highly data efficient. It is worth noting that our method can still reconstruct the scene with reasonable accuracy when $\tau = 1000 \mathit{ms}$, where only 3 equivalent views are used and each pixel can only generate at most 4 events throughout the $4000\mathit{ms}$ sequence. The event stream is also around $200 \times$ sparser than the default with $\tau = 0 \mathit{ms}$.

\clearpage

\begin{table*}[t!]
% \small
\setlength{\tabcolsep}{13.7pt}

\begin{center}
\begin{tabular}{rccclccc}
\toprule
\multirow{2}{*}{$\tau$, $\mathit{ms}$} & \multicolumn{3}{c}{Statistics} & \hspace{-1em} & \multicolumn{3}{c}{Metrics} \\ \cmidrule(lr){2-4} \cmidrule(l){6-8} 
 & \% Seq. Duration & Sparsity, $\times$ & Equiv. \# Views & \hspace{-1em} & PSNR $\uparrow$ & SSIM $\uparrow$ & LPIPS $\downarrow$ \\ \midrule
0 & 0 & 1.000 & 336.8 & \hspace{-1em} & 30.24 & 0.958 & 0.040 \\
8 & 0.2 & 4.176 & 80.66 & \hspace{-1em} & 30.41 & 0.959 & 0.042 \\
25 & 0.625 & 8.440 & 39.90 & \hspace{-1em} & 29.84 & 0.958 & 0.041 \\
50 & 1.25 & 13.50 & 24.95 & \hspace{-1em} & 29.20 & 0.953 & 0.046 \\
100 & 2.5 & 21.27 & 15.83 & \hspace{-1em} & 27.40 & 0.938 & 0.060 \\
250 & 6.25 & 40.80 & 8.255 & \hspace{-1em} & 25.95 & 0.916 & 0.081 \\
500 & 12.5 & 67.77 & 4.970 & \hspace{-1em} & 24.08 & 0.900 & 0.102 \\
1000 & 25 & 110.5 & 3.048 & \hspace{-1em} & 22.10 & 0.854 & 0.204 \\
2000 & 50 & 209.6 & 1.607 & \hspace{-1em} & 17.05 & 0.762 & 0.398 \\ \bottomrule
\end{tabular}
\end{center}

\caption{Robustness of our method to temporal event sparsity on the \texttt{chair} scene.}
\label{tab:efficiency}
\end{table*}

\end{document}